\documentclass{article}
\usepackage{amsmath,graphicx,bm,stmaryrd,amsfonts,amsmath,algorithm,color,caption,subcaption,authblk}

\twocolumn
\def\ninept{\def\baselinestretch{.95}\let\normalsize\small\normalsize}

\def\@maketitle{\newpage
 \null
 \vskip 3em \begin{center}
 {\large \bf \@title \par} \vskip 1.5em {\large \lineskip .5em
\begin{tabular}[t]{c}\@name \\ \@address
 \end{tabular}\par} \end{center}
 \par
 \vskip 1.5em}

\usepackage[noend]{algpseudocode}

\usepackage{amsmath}

\usepackage{amsfonts}

\usepackage{amssymb}

\usepackage{stmaryrd}

\usepackage[mathscr]{eucal}

\usepackage[normalem]{ulem}

\usepackage{amsbsy}

\usepackage{bm}

\usepackage{bbding}

\usepackage{array}

\usepackage{tabularx}

\usepackage[neveradjust]{paralist}

\usepackage{fancyvrb}

\usepackage{caption}

\usepackage{placeins}

\usepackage{ifpdf}
\ifpdf
\DeclareGraphicsExtensions{.pdf,.png}
\else
\DeclareGraphicsExtensions{.eps}
\fi

\newcommand{\Sec}[1]{\hyperref[sec:#1]{\S\ref*{sec:#1}}} 
\newcommand{\Section}[1]{\hyperref[sec:#1]{Section~\ref*{sec:#1}}} 
\newcommand{\AppFull}[1]{\hyperref[sec:#1]{Appendix~\ref*{sec:#1}}} 
\newcommand{\Eqn}[1]{\hyperref[eq:#1]{(\ref*{eq:#1})}} 
\newcommand{\Fig}[1]{\hyperref[fig:#1]{Figure~\ref*{fig:#1}}} 
\newcommand{\Tab}[1]{\hyperref[tab:#1]{Table~\ref*{tab:#1}}} 
\newcommand{\Thm}[1]{\hyperref[thm:#1]{Theorem~\ref*{thm:#1}}} 
\newcommand{\Cor}[1]{\hyperref[cor:#1]{Corollary~\ref*{cor:#1}}} 
\newcommand{\Alg}[1]{\hyperref[alg:#1]{Algorithm~\ref*{alg:#1}}} 
\newcommand{\Def}[1]{\hyperref[def:#1]{Definition~\ref*{def:#1}}} 

\newcommand{\Real}{{\mathbb R}}

\newcommand{\Tra}{^{{\sf T}}} 
\newcommand{\V}[1]{{\bm{\mathbf{\MakeLowercase{#1}}}}} 
\newcommand{\Vhat}[1]{{\bm \hat{\mathbf{\MakeLowercase{#1}}}}} 

\newcommand{\M}[1]{{\bm{\mathbf{\MakeUppercase{#1}}}}} 
\newcommand{\MC}[2]{\V{#1}_{#2}} 
\newcommand{\MhatC}[2]{\Vhat{#1}_{#2}} 
\newcommand{\Mn}[2]{\M{#1}^{(#2)}} 

\newcommand{\T}[1]{\boldsymbol{\mathscr{\MakeUppercase{#1}}}} 

\newcommand{\fnorm}[1]{\left\lVert \, #1 \, \right\rVert_{F}}

\let\oldthebibliography\thebibliography
\renewcommand\thebibliography[1]{%
  \oldthebibliography{#1}%
  \setlength{\itemsep}{0pt plus 0.3ex}%
}

\usepackage{mathtools}
\mathtoolsset{showonlyrefs}

\title{dCMF: Learning interpretable evolving patterns from temporal multiway data}

\author[1,2]{Christos Chatzis}
\author[1]{Carla Schenker}
\author[3]{Jérémy E. Cohen}
\author[1]{Evrim Acar}
\affil[1]{Simula Metropolitan Center for Digital Engineering, Oslo, Norway}
\affil[2]{Faculty of Technology, Art and Design, OsloMet, Oslo, Norway}
\affil[3]{Univ Lyon, INSA-Lyon, UCBL, UJM, CNRS, Lyon, France}
\date{}
\begin{document}

\maketitle

\begin{abstract}
Multiway datasets are commonly analyzed using unsupervised matrix and tensor factorization methods to reveal underlying patterns. Frequently, such datasets include timestamps and could correspond to, for example, health-related measurements of subjects collected over time. The temporal dimension is inherently different from the other dimensions, requiring methods that account for its intrinsic properties. Linear Dynamical Systems (LDS) are specifically designed to capture sequential dependencies in the observed data. In this work, we bridge the gap between tensor factorizations and dynamical modeling by exploring the relationship between LDS, Coupled Matrix Factorizations (CMF) and the PARAFAC2 model. We propose a time-aware coupled factorization model called d(ynamical)CMF that constrains the temporal evolution of the latent factors to adhere to a specific LDS structure. Using synthetic datasets, we compare the performance of dCMF with PARAFAC2 and t(emporal)PARAFAC2 which incorporates temporal smoothness. Our results show that dCMF and PARAFAC2-based approaches perform similarly when capturing smoothly evolving patterns that adhere to the PARAFAC2 structure. However, dCMF outperforms alternatives when the patterns evolve smoothly but deviate from the PARAFAC2 structure. Furthermore, we demonstrate that the proposed dCMF method enables to capture more complex dynamics when additional prior information about the temporal evolution is incorporated.
\end{abstract}

\section{Introduction}

With the rapid growth in the availability of measurements in recent years, many datasets now inherently exhibit multiway structure, allowing for variations along multiple dimensions. The analysis of such data, a challenge that arises in numerous domains, has been effectively tackled using tensor factorizations (e.g. \cite{KoBa09,AcYe09}), i.e., extensions of matrix factorizations to multiway arrays. In their essence, these methods often enable the decomposition of the data into interpretable factors that reveal the underlying patterns.

\begin{figure}[tb]
\centering
\includegraphics[width=\columnwidth]{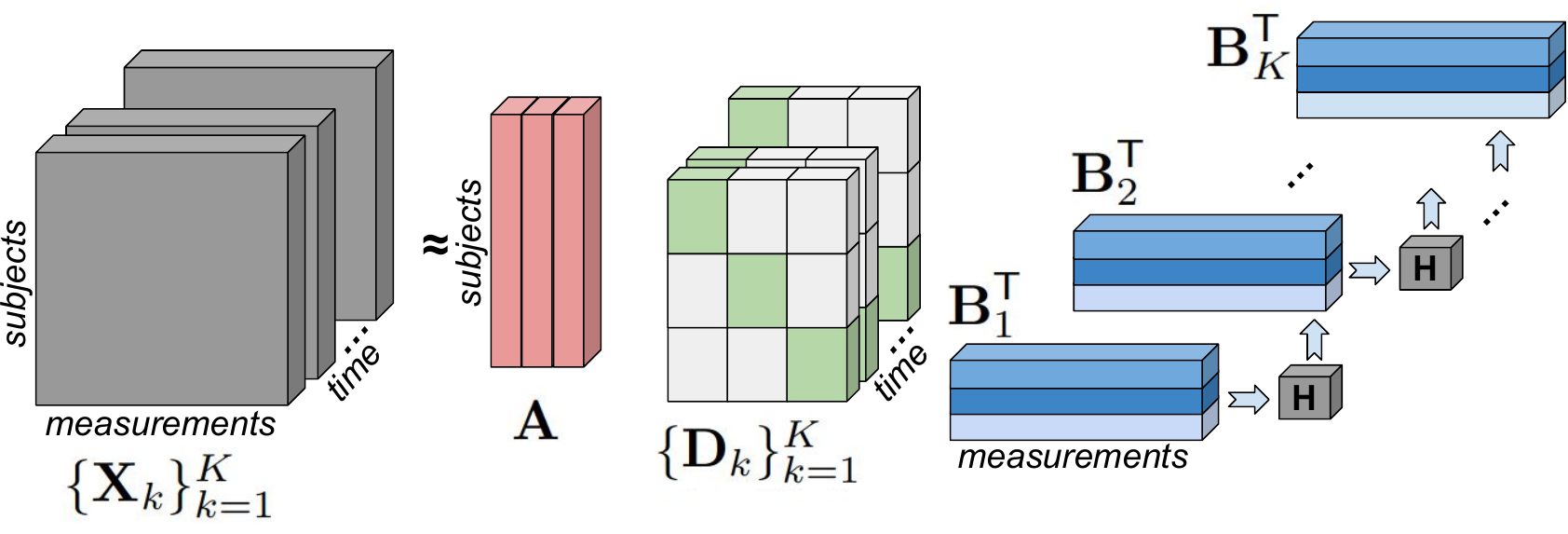}
\caption{The dCMF model. Evolving patterns captured by the columns of factor matrix $\M{B}_k$ are related to the previous $\M{B}_{k-1}$ through the transition matrix $\M{H}$.
}
\label{dcmf_illustration}
\end{figure}

Frequently, one of the ``ways" that the data evolves across is time, implying that patterns can change over time. Consequently, the objective of such analyses often becomes the identification and tracking of these dynamics. Standard methods can be used off-the-shelf to analyze such temporal data. For example, the CANDECOMP/PARAFAC (CP) tensor model \cite{parafac} can be used to extract a temporal factor that captures the evolution of pattern strength over time. However, the underlying structure remains fixed. PARAFAC2 \cite{parafac2} addresses this limitation by allowing factors that capture structural changes, though specific structural constraints must hold. In contrast, coupled matrix factorizations (CMF) \cite{cmf} impose no such constraints, offering greater flexibility but lacking uniqueness guarantees. These approaches, however, do not account for the inherent sequential nature of the time dimension. Taking the temporal aspect into account offers several advantages, including improved user recommendations \cite{timesvd}, more accurate future predictions \cite{LiHa16}, and robustness to noise \cite{tparafac2,tparafac2_2}. As a result, there are numerous approaches incorporating time in the literature. A prevalent strategy is temporal regularization. Some approaches impose such regularization on the strength of the uncovered patterns over time: Temporal Regularized Matrix Factorization (TRMF) \cite{trmf} assumes an autoregressive structure and ATOM \cite{atom} assumes smoothness. Others focus on the structure of latent factors such as Sun et al. \cite{assistants} integrating Kalman filter-based temporal regularization. Similarly, Appel et al. \cite{ApCuAg18} assume smoothness on the evolution (of latent structures) of a coupled CMF model while t(emporal)PARAFAC2 \cite{tparafac2} promotes smoothly changing factors. Apart from temporal regularization, other approaches include enforcing smoothness through small gradient steps on factor updates when new data is received \cite{KaBh21} and modeling one of the factors of a CMF as a function of time \cite{YuAgWa17}.

Another approach to studying evolving patterns are Linear Dynamical Systems (LDS). For instance, many studies in hyperspectral imaging utilize LDS and learn evolving structures from raw data \cite{HeChJu16,10103834}. 
Closer to our work is the g(eneralized)LDS framework \cite{LiHa16}, which receives as input (a matrix formed from) multiple multivariate time series and factorizes the input into a common ``emission" matrix and a factor corresponding to hidden states of an LDS. Computing the transition matrix of this system is part of the optimization procedure. While our work does not currently involve learning the transition matrix, it extends gLDS by introducing support for tensor inputs and enabling latent factors to evolve independently and under different dynamical systems.

Analyzing temporal multiway data with the goal of capturing interpretable evolving patterns requires methods with specific properties. First, \textbf{time-awareness} is essential, as the approach must explicitly account for the temporal dimension. Second, the method should allow \textbf{sufficient structural flexibility} in the latent factors. Imposing overly rigid constraints can lead to inaccurate factor estimates. Third, the \textbf{uniqueness} of the factorization is imperative for interpretability, ensuring that the extracted factors correspond to identifiable components. To the best of our knowledge, there is a lack of methods fulfilling these properties. 

In this work, we propose d(ynamical)CMF, a coupled factorization approach that involves temporal regularization based on LDS (Figure \ref{dcmf_illustration}). The framework enables incorporating prior information on the factors by choosing a-priori the transition matrix {\small $\M{H}$} for the LDS. Our contributions are as follows:
\begin{itemize}
    \item We introduce the dCMF model and demonstrate that incorporating the proposed LDS penalty results in a time-aware coupled factorization. We also explore the relation between dCMF and CP, PARAFAC2 models as well as temporal smoothness.
    \item With experiments on synthetic data, we demonstrate that dCMF is accurate and flexible since it can capture the evolving patterns that PARAFAC2-based methods would not. Further, the uniqueness of the solution is empirically discussed in different experimental settings.
    \item We also demonstrate that if accurate prior information is incorporated in the dCMF model, high accuracy can be expected even in settings with high noise levels.
\end{itemize}

\section{Background}

Let {\small$\T{X} \in \Real^{I \times J \times K}$} denote a third-order tensor, where the last mode ({\small$K$}-sized) indexes the time points. Such data can be analyzed by a coupled matrix factorization approach \cite{cmf} with each frontal slice {\small $\M{X}_k \in \Real^{I \times J}$} modelled as:

{\small 
\begin{equation} \label{eq:cmf}
\M{X}_k \approx \M{A} \M{B}_k\Tra \quad \forall k \leq K 
\end{equation}}where {\small $\M{A}\in\Real^{I \times R}$} and {\small $\M{B}_k\in\Real^{J \times R} \ \forall k \leq K$} with {\small$R$} denoting the number of components. Interpreting the columns of the factors yields a summary of the main patterns present, which can change across {\small $k$}. However, there are no uniqueness guarantees when solving for the factorization unless additional constraints (e.g. non-negativity) are imposed. The PARAFAC2 model \cite{parafac2}, which can be considered a CMF approach, utilizes an additional factor {\small $\M{C}$} and is formulated as:

{\small
\begin{align}
    \begin{split} 
        \M{X}_k & \approx \M{A} \M{D}_k \M{B}_k\Tra \\
        \M{B}_k & = \M{P}_k \M{B}
    \end{split} \quad \forall k \leq K \label{eq:parafac2}
\end{align}}where {\small $\M{A}\in\Real^{I \times R}$}, {\small $\M{C}\in\Real^{K \times R} \ \forall k \leq K$}, {\small$\M{D}_k = diag(\M{C}_{k,:}) \ \forall k \leq K$}, {\small $\M{P}_k \in \Real^{J \times R} \ \forall k \leq K$} have orthonormal columns and {\small $\M{B} \in \Real^{R \times R}$}. The second line is necessary for uniqueness of the patterns (up to permutation and scaling ambiguities)\cite{parafac2_direct} and referred to as the PARAFAC2 constraint. Roald et al. \cite{parafac2_evolution} demonstrated the model's ability to capture evolving patterns. Nevertheless, when analyzing temporal data, this method suffers from two drawbacks: (a) it does not take into account the sequential nature of the temporal dimension, and (b) the uncovered latent factors must adhere to the PARAFAC2 constraint. To tackle (a), t(emporal)PARAFAC2 \cite{tparafac2,tparafac2_2} was proposed based on the following optimization problem:

{\small
\begin{equation} \label{tparafac2}
\begin{split} 
     & \min_{\M{A},\{\M{B}_k\}_{k=1}^K,\M{C}}  \quad \Bigg\{ \sum_{k=1}^{K} \fnorm{\M{X}_k - \M{A} \M{D}_k \M{B}_k^T}^2 + g_{\M{A}}(\M{A})    \\ 
    & + \lambda_B \sum_{k=1}^{K-1} \fnorm{ \M{B}_{k+1} - \M{B}_{k} }^{2} + g_{\M{D}}(\{\M{D}_k\}_{k=1}^{K}) \Bigg\} \\
    & \ \ \text{subject to} \ \  \quad \M{B}_k = \M{P}_k \M{B} \quad \forall  k \leq K
\end{split}
\end{equation}}where smooth changes across the evolving factors {\small $\M{B}_k$} are assumed and {\small $g_{\M{A}}$} and {\small $g_{\M{D}}$} indicate norm-based penalties on the respective factors (necessary due to scaling ambiguity). However, the issue in (b) persists, potentially distorting the uncovered latent structures when data does not follow the PARAFAC2 constraint. tPARAFAC2 also struggles to capture more complex temporal dynamics beyond smooth changes.

\section{Proposed method: \lowercase{d(ynamical)}CMF}
To address these points, we propose the d(ynamical)CMF model, which is a CMF-based approach possessing temporal regularization based on LDS instead of the PARAFAC2 constraint. We discuss how the proposed approach is related to CP, PARAFAC2 and tPARAFAC2 here, and compare them using numerical experiments in Section \ref{sec:Experiments}. In Section \ref{sec:uniq}, we discuss the uniqueness of the dCMF solution empirically.

\subsection{The LDS constraint}
In dCMF, we regularize each component of the evolving factors to adhere to the structure of an LDS, i.e.
\addtocounter{equation}{1}

{\small
\begin{align} \label{lds-formulation-full}
        \begin{split}
        \M{X}_k & \approx \M{A} \M{D}_k \M{B}_k\Tra \  \forall k \leq K \\
        \V{b}_{k+1,r} & \approx \Mn{H}{r}  \V{b}_{k,r} \  \forall k < K \ , \ r \leq R
    \end{split} 
\end{align}}with {\small $\V{b}_{k,r}=\M{B}_k(:,r)$} and {\small$\{\Mn{H}{r}\}_{r=1}^{R}$} corresponding to the transition matrices of the systems. 

If {\small $\Mn{H}{1}=\Mn{H}{2}=\dots=\M{H}=\M{I}$} and the relation between temporal factors holds exactly, Equation \eqref{lds-formulation-full} corresponds to the CP model since {\small $\M{B}_1 = \M{B}_2 = \dots = \M{B}_K = \M{B}$}. If, more generally, we assume that all transition matrices are equal to an orthogonal matrix {\small$\M{H}$}, then this adheres to the PARAFAC2 constraint, which facilitates uniqueness. A specific instance of this would be setting {\small$\M{H}=\M{I}$}, where temporal smoothness similar to \eqref{tparafac2} is assumed. Furthermore, even if we relax the orthogonality requirement for {\small $\M{H}$}, we still obtain a time-aware factorization of the input, i.e., in PARAFAC2 the factors that correspond to the {\small $k$}-th and {\small $l$}-th time-point (w.l.o.g. {\small $k < l$}) are {\small$\M{B}_l=\M{P}_l \M{B}$} and {\small$\M{B}_k=\M{P}_k \M{B}$} and related through {\small $\M{B}_l = \M{P}_l \M{P}_k\Tra \M{B}_k$} irrespective of their order in time whereas using the LDS constraint, the relation becomes {\small $\M{B}_l = \M{H}^{l-k} \M{B}_k$}, which takes into account the time points of the observations. We restrict our analysis to LDS regularization using a single, potentially non-orthogonal, transition matrix as follows:

{\small
\begin{align} \label{lds-formulation}
        \begin{split}
        \M{X}_k & \approx \M{A} \M{D}_k \M{B}_k\Tra \  \forall k \leq K \\
        \V{b}_{k+1,r} & \approx \M{H}  \V{b}_{k,r} \  \forall k < K \ , \ r \leq R \ .
    \end{split} 
\end{align}}

\subsection{Optimization}

We formulate \eqref{lds-formulation} as the following optimization problem:

{\small
\begin{equation} \label{dcmf}
\begin{split} 
     & \min_{\M{A},\{\M{B}_k\}_{k=1}^K,\M{C}} \quad  \Bigg\{ \sum_{k=1}^{K} \fnorm{\M{X}_k \!-\! \M{A} \M{D}_k \M{B}_k^T}^2 + \lambda \fnorm{\M{A}}^2    \\ 
      & + \lambda_B \sum_{k=1}^{K-1}\fnorm{ \M{B}_{k+1} \!-\! \M{H} \M{B}_{k} }^{2} + \sum_{k=1}^{K} \bigl( \lambda \fnorm{\M{D}_k}^2 + \iota_{\mathbb{R}_{+}}(\M{D}_k)  \bigr) \Bigg\}
\end{split}
\end{equation}}where the first and third terms denote the fidelity term and proposed LDS constraint, respectively. Due to the scaling ambiguity, it is necessary to impose ridge regularization on factors {\small $\M{A}$} and {\small $\{\M{D}_k\}_{k=1}^{K}$} \cite{parafac2-aoadmm}, but other types of norm-based regularization are applicable as well. The LDS constraint is included as a soft regularization penalty, which gives the method more structural freedom (see Section \ref{sec:non_par2}). We overcome the sign ambiguity by imposing non-negativity on {\small $\{\M{D}_k\}_{k=1}^{K}$}, with {\small $\iota_{\mathbb{R}_{+}}$} denoting the indicator function of the non-negative orthant.


We solve this optimization problem using an Alternating Optimization - Alternating Direction Method
of Multipliers (AO-ADMM)-based algorithm \cite{aoadmm,parafac2-aoadmm}. This algorithmic framework is highly flexible, allowing for the straightforward inclusion of various types of regularization in all modes to accommodate different application requirements. We form the augmented Lagrangian of \eqref{dcmf} as follows:

{\small \begin{align} 
    &\mathcal{L} =\sum_{k=1}^{K}\lVert \M{X}_k - \M{A}\M{D}_k\M{B}_k\Tra\rVert^2_{F} + \lambda \lVert \M{A} \rVert^{2}_F + \lambda \sum_{k=1}^K \lVert \M{D}_k \rVert^{2}_F \notag  \\
    &+ \lambda_B \sum_{k=2}^K\lVert \M{Z}_{\M{B}_k}- \M{H} \M{Z}_{\M{B}_{k-1}} \rVert^2_F + \sum_{k=1}^K \frac{\rho_{\M{B}_k}}{2}\lVert \M{B}_k -  \M{Z}_{\M{B}_k} + \M{\mu}_{\bm{B}_k} \rVert^2_{F} \notag \\
    &+ \sum_{k=1}^K\iota_{\mathbb{R}_{+}}(\M{Z}_{\M{D}_{k}})  + \sum_{k=1}^K \frac{\rho_{\M{D}_{k}}}{2}\lVert  \M{D}_k  - \M{Z}_{\M{D}_{k}}  + \M{\mu}_{\M{D}_{k}} \rVert^2_{F} ,
    \label{eq:lagrangian}
\end{align}}where the auxiliary variables {\small $\{\M{Z}_{\M{B}_k},\M{Z}_{\M{D}_k}\}_{k=1}^{K}$}, dual variables {\small $\{\M{\mu}_{\M{B}_k},\M{\mu}_{\M{D}_k}\}_{k=1}^{K}$} and feasibility penalties {\small $\{\rho_{\M{B}_k},\rho_{\M{D}_k}\}_{k=1}^{K}$} have been introduced. We then iteratively solve the subproblems for each mode until convergence is achieved (small relative or absolute change in function value of \eqref{dcmf}) or a pre-set maximum number of iterations is reached. Our algorithm follows \cite{parafac2-aoadmm}, with the addition of the update of {\small $\{\M{Z}_{B_k}\}_{k=1}^{K}$}, for which a block diagonal system has to be solved.

\section{Experimental evaluation}
\label{sec:Experiments}

\begin{figure*}[htbp]
\centering
\includegraphics[width=\textwidth]{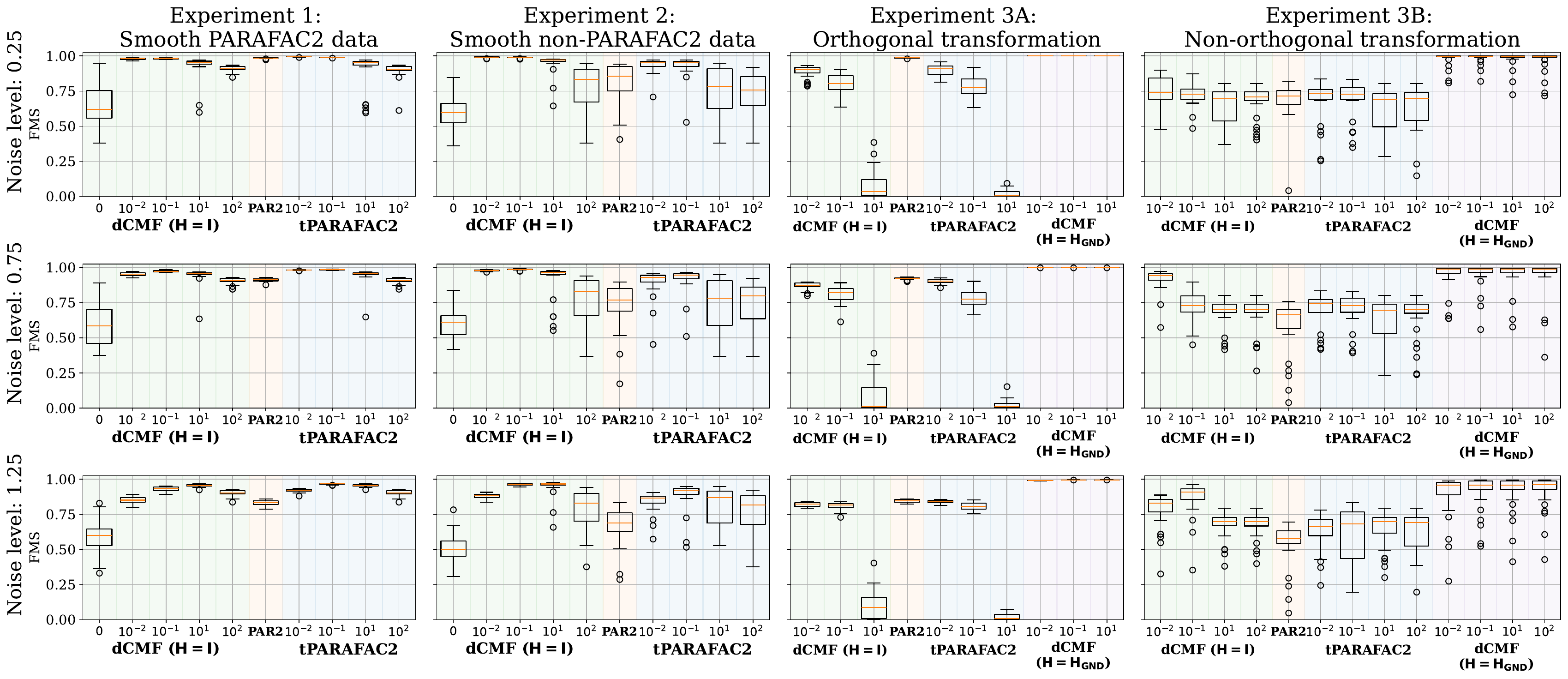}
\caption{Accuracy of methods in terms of FMS in each experiment. Each boxplot contains 30 points, one for the best run of each method for each dataset. Multiple regularization strengths are considered for temporal smoothness (tPARAFAC2) and the proposed LDS constraint (dCMF). PAR2 is short for PARAFAC2.}
\label{results}
\end{figure*}


\subsection{Experimental setup}

We evaluate the performance of the proposed method using synthetic datasets, each with dimensions {\small $30 \times 40 \times 35$}. To assess the accuracy of the method in terms of finding the underlying patterns under different scenarios, we generate four types of synthetic data, with details provided in the subsequent subsections. In all cases, we generate the ground truth factors {\small $\M{A},\{\M{B}_k\}_{k=1}^K,\M{C}$} with {\small $R=3$} components, form each slice of the input data {\small$\M{X}_k = \M{A} \M{D}_k \M{B}_k\Tra \ \forall k \leq K$} and add noise according to

{\small
\begin{equation*}
    \T{X}_{noisy} = \T{X} + \eta \fnorm{ \T{X} }\frac{\Theta}{\fnorm{ \Theta}}
\end{equation*}}where {\small $\Theta \sim \mathcal{N}(0,1)$} with {\small $\eta=0.25, 0.75, 1.25$} denoting the noise level. The goal is to recover the underlying factors from noisy data. We assess accuracy using the Factor Match Score (FMS):

{\small
\begin{equation*}
    \text{FMS} \small{= \sum_{i=1}^R\frac{|\MhatC{A}{i}\Tra\MC{A}{i}|}{\lVert \MhatC{A}{i} \rVert \lVert \MC{A}{i} \rVert} \frac{|\MhatC{B}{i}\Tra\MC{B}{i}|}{\lVert \MhatC{B}{i} \rVert \lVert \MC{B}{i} \rVert}  \frac{|\MhatC{C}{i}\Tra\MC{C}{i}|}{\lVert \MhatC{C}{i} \rVert \lVert \MC{C}{i} \rVert}},
\end{equation*}}where {\small $\MC{A}{i}$}, {\small $\MC{B}{i}$}, and {\small $\MC{C}{i}$} represent the {\small $i$}-th column of the true factors, and {\small $\MhatC{A}{i}$}, {\small $\MhatC{B}{i}$}, and {\small $\MhatC{C}{i}$} are the factors estimated by the model (after finding the optimal permutation). {\small $\MhatC{B}{i}$} and {\small $\MC{B}{i}$} are formed by stacking the {\small $i$}-th column across all {\small$\{\M{B}_k\}_{k=1}^{K}$}.

Our implementation\footnote{https://github.com/cchatzis/temp-dcmf-repo} builds on \cite{tensorly,matcouply,tlviz}. AO-ADMM is used for all methods. For all experiments, the maximum number of iterations was set to {\small $8000$}, the outer relative and absolute tolerance to {\small $10^{-8}$} and the inner relative tolerance to {\small $10^{-5}$}. The feasibility criterion for all constraints was set to {\small $10^{-5}$}. Twenty initializations were used for each dataset and all runs that resulted in solutions with feasibility gaps larger than {\small $10^{-5}$} and degenerate solutions \cite{KoBa09} were discarded from the analysis. Out of the remaining runs for each dataset, we designate as `best' the run with the lowest function value.


\subsection{Smooth PARAFAC2 data} \label{sec:par2}

In this experiment, we compare the performance of PARAFAC2, tPARAFAC2, and the proposed dCMF on smooth data exhibiting PARAFAC2 structure. For dCMF, we set {\small$\M{H}=\M{I}$}, since the data is smoothly changing. We generate 30 datasets, where the ground truth factors are drawn as {\small $\M{A} \sim \mathcal{N}(0, 1)$} and {\small $\M{C} \sim \mathcal{U}(1, 15)$}. For the evolving mode, we designate specific indices as ‘active’ for each pattern, setting all remaining entries to zero to ensure orthogonality between patterns. We draw the active entries of {\small$\M{B}_1$} from {\small$\mathcal{N}(0,1)$} and incrementally add values from {\small $\mathcal{N}(0,0.25)$} to active indices of each pattern to obtain {\small $\M{B}_2, \M{B}_3, ..., \M{B}_K$}. We then column-normalize each {\small$\M{B}_k$}. With this process, we make sure the ground truth has PARAFAC2 structure since (a) the off-diagonal entries of the cross product {\small $\M{B}_k\Tra \M{B}_k$} are constantly zero due to the patterns not overlapping and (b) the diagonal entries are constant since the evolving factors are column-normalized. Here, we expect PARAFAC2-based methods to perform well.

The results of this experiment are shown in the first column of Figure \ref{results}, where various regularization strengths {\small $\lambda_B$} are considered. Setting {\small $\lambda_B=0$} for dCMF results in an unconstrained CMF model (not considering ridge penalties), yielding non-unique output. As the assumption that the data is smooth holds, increasing {\small $\lambda_B$} results in more accurate recovery for both dCMF and tPARAFAC2, especially for higher noise levels, with tPARAFAC2 having slightly higher accuracy.

\subsection{Non-PARAFAC2 smooth data} \label{sec:non_par2}

This setting investigates the performance of PARAFAC2, tPARAFAC2, and dCMF (with {\small $\M{H}=\M{I}$}) on synthetic data that significantly violates the PARAFAC2 constraint. 30 datasets are randomly generated with {\small $\M{A} \sim \mathcal{N}(0, 1)$} and {\small $\M{C} \sim \mathcal{U}(1, 15)$}. According to the PARAFAC2 constraint, {\small $\M{B}_k\Tra \M{B}_k$} has to remain constant {\small $\forall k \leq K$}. For a single time point {\small $k$}, off-diagonal entries of the cross product matrix capture the cosine similarity of evolving patterns (i.e., {\small $\M{B}_{k}(:,i) \ \text{and} \ \M{B}_{k}(:,j)$}) after column normalization which has to, according to the PARAFAC2 constraint, remain constant. To violate this constraint, we create datasets by changing the similarity between the components across time by introducing structured steps from one pattern towards (or away from) another. To instantiate, component 1 will be taking steps towards component 2 for each {\small $k\in \mathcal{R}_1 = [m, n]_{\mathbb{Z}}, 1 \leq m \leq n \leq K$} and steps ``away" for the rest of the time points:

{\small
\begin{equation*}
\begin{split}
    \M{B}_k \gets \underbrace{\M{B}_{k-1} + \mathcal{N}(0,0.1)}_{\text{smooth changes}} + \M{\Delta}_k ,\qquad \qquad \\
  \M{\Delta}_k(:, 1) = 
   \begin{cases}
    \text{scale}_{1,k} \! \cdot \! (\M{B}_{k}(:, 2) - \M{B}_{k}(:, 1)) & \text{if } k \in \mathcal{R}_1, \\
   -1 \! \cdot \! \text{scale}_{1,k} \! \cdot \! (\M{B}_{k}(:, 2) - \M{B}_{k}(:, 1)) & \text{otherwise},
\end{cases}
\end{split}
\end{equation*}}where {\small $\text{scale}_{1,k}\sim\mathcal{U}(0.04,0.08) \ \forall k \leq K$} is the step size. A similar scheme is used for the rest of the components. Steps are small and the data remains smooth. See the GitHub repo for more details on data generation.

The second column of Figure \ref{results} presents the results. Since the data is smooth, increasing regularization helps mitigate noise. However, PARAFAC2-based methods struggle more to recover the ground truth without strong regularization due to enforced constraint. In contrast, dCMF shows better performance due to its greater structural flexibility.

\subsection{Non-identity transition matrix}

Here, the goal is to investigate whether incorporating prior knowledge via {\small$ \M{H}$} in dCMF increases accuracy. We consider two settings, i.e.,  one with an orthogonal and another with a non-orthogonal transition matrix {\small$\M{H}$}.

In the first setting ({orthogonal \small$\M{H}$}), thirty datasets are created, with {\small $\M{A} \sim \mathcal{N}(0, 1)$} and {\small $\M{C} \sim \mathcal{U}(1, 15)$}. For each evolving pattern, certain consecutive indices are chosen to be active and we generate the active indices of (the orthogonal) {\small $\M{B}_1 \sim \mathcal{N}(0, 1)$}. 
Three random orthogonal transition matrices {\small $\Mn{H}{1}, \Mn{H}{2}, \Mn{H}{3}$} are generated, each describing the evolution of the structure of the respective pattern. These matrices are combined to form the block diagonal matrix {\small $\M{H} = \text{diag}(\Mn{H}{1}, \Mn{H}{2}, \Mn{H}{3})$}. Then, we compute {\small $\M{B}_{k+1} = \M{H} \M{B}_{k} \ \forall k < K$}, with each {\small $\Mn{H}{r}$} only applied to the active indices of the {\small$r$}-th column of {\small $\M{B}_{k}$}. Here, {\small$\M{H}$} is used as prior information ({\small $\M{H}_{\text{GND}}=\M{H}$}) when fitting the dCMF model.

In the second setting (non-orthogonal {\small$\M{H}$}), thirty more datasets are generated but instead, a shared non-orthogonal transition matrix is used for all components. After sampling {\small $\M{B}_1 \sim \mathcal{N}(0,1)$}, we generate {\small $\M{H}=\M{V}\M{\Lambda}\M{V}^{-1}$} with {\small $\M{V}\sim\mathcal{N}(0,1)$} and {\small $diag(\M{\Lambda})\sim\mathcal{N}(0.95,0.1)$} and construct {\small $\M{B}_{k+1}=\M{H}\M{B}_{k} \ \forall k < K$}. {\small $\M{A}$} and {\small $\M{C}$} are generated similarly to the previous setting. Since the transition matrix is non-orthogonal, the ground truth does not satisfy the PARAFAC2 constraint. Nevertheless, the goal remains to investigate if incorporating prior knowledge through the transition matrix increases recovery accuracy.

We compare the accuracy of the recovered factors from PARAFAC2, tPARAFAC2 and dCMF with {\small $\M{H}=\M{I}$} and the ground truth {\small $\M{H}_{\text{GND}}$} of each dataset. The results are shown in the third and fourth columns of Figure \ref{results}. For the orthogonal transition matrix case, we notice that since the data is not smooth, involving smoothness either with the LDS constraint or tPARAFAC2 does not result in higher accuracy. Instead, we see PARAFAC2 performing better than tPARAFAC2 and dCMF. However, if the appropriate prior information is used, dCMF always perfectly recovers the ground truth ({\small $\text{FMS}>0.99$}) even in the most noisy setting. For the non-orthogonal transition matrix case, the data is smooth ({\small $\M{H}_{\text{GND}}$} is close to identity by definition), and therefore temporal smoothness improves recovery. No dataset possesses PARAFAC2 structure, hence the decreased performance of PARAFAC2-based methods while the best results are achieved by dCMF ({\small $\M{H}=\M{H}_{\text{GND}}$}).

\subsection{Uniqueness} \label{sec:uniq}

\begin{figure}[tb]
\centering
\includegraphics[width=\columnwidth]{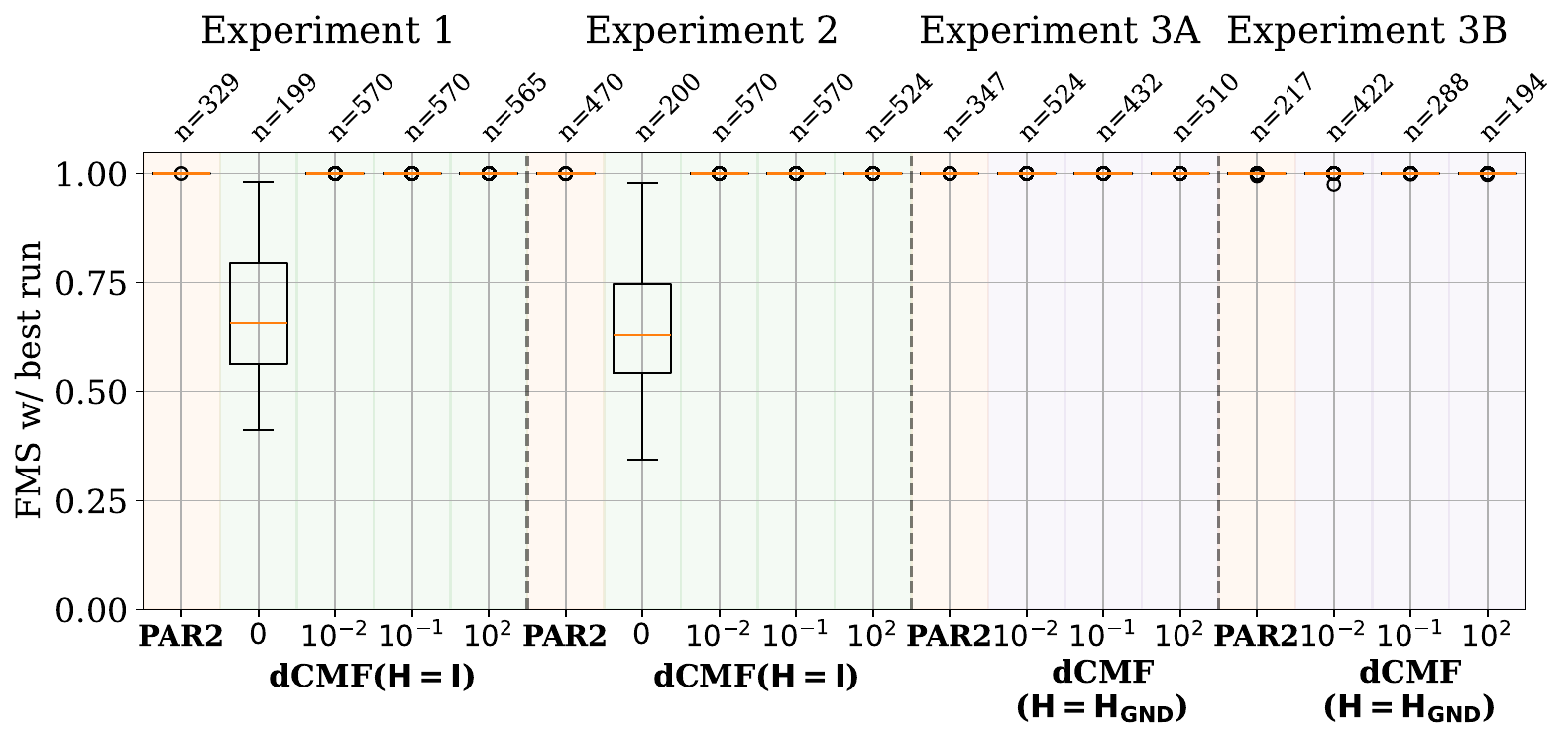}
\caption{Boxplots show the FMS of all runs when compared with the best run for each dataset (at noise level {\small $\eta=0.75$}). For each experiment, there are in total 570 runs (30 datasets $\times$ 20 initializations $-$ the best run for each dataset). {\small$n$} denotes the number of runs after discarding failed runs.}
\label{uniqueness}
\end{figure}

Uniqueness is critical for the interpretability of the patterns extracted using factorization methods. We empirically investigate the uniqueness of dCMF by comparing all (non-failed) runs with the best run for each dataset. We consider a run as `failed' if the solution was infeasible, exited due to numerical issues, was degenerate or had a function value with less than 6 decimal digits shared with the best run (to discard local minima). In Figure \ref{uniqueness}, where we also include PARAFAC2 for reference, we can observe that introducing the proposed LDS constraint (with appropriate strength) helps to achieve a unique solution. The results are similar for different noise levels.

\section{Conclusion}

In this paper, we have introduced a soft constraint for coupled matrix factorizations that regularizes evolving factors to promote the structure of a linear dynamical system. Setting the transition matrix to identity, in particular, induces temporal smoothness. Experimental results indicate that (a) dCMF can capture dynamics that do not conform to PARAFAC2 better than PARAFAC2-based methods and (b) if the transition matrix of the evolving factors is known a-priori, incorporating it into the dCMF framework enables highly accurate recovery of the underlying patterns. As future work, we plan to focus on computational efficiency and learning the transition matrix.

\bibliography{bibliography}

\end{document}